\documentclass{article}

\usepackage{times}
\usepackage{graphicx}
\usepackage{subfigure}
\usepackage{amsmath, amsfonts, bm, mathrsfs}
\usepackage{float}

\usepackage{natbib}

\usepackage{algorithm}
\usepackage{algorithmic}
\usepackage{hyperref}

\newcommand{\iid}{\overset{i.i.d.}{\sim}}
\newcommand{\argmin}{\arg\min}

\usepackage[accepted]{icml2016mod}

\icmltitlerunning{DR-ABC: Approximate Bayesian Computation with Kernel-Based Distribution Regression}

\begin{document}

\twocolumn[
\icmltitle{DR-ABC: Approximate Bayesian Computation with \\ Kernel-Based Distribution Regression}

\icmlauthor{Jovana Mitrovic}{mitrovic@stats.ox.ac.uk}
\icmlauthor{Dino Sejdinovic}{dino.sejdinovic@stats.ox.ac.uk}
\icmlauthor{Yee Whye Teh}{y.w.teh@stats.ox.ac.uk}
\icmladdress{Department of Statistics, University of Oxford}

\icmlkeywords{Approximate Bayesian Computation, Kernel methods}

\vskip 0.1in
]

\begin{abstract}
Performing exact posterior inference in complex generative models is often difficult or impossible due to an expensive to evaluate or intractable likelihood function. Approximate Bayesian computation (ABC) is an inference framework that constructs an approximation to the true likelihood based on the similarity between the observed and simulated data as measured by a predefined set of summary statistics. Although the choice of appropriate problem-specific summary statistics crucially influences the quality of the likelihood approximation and hence also the quality of the posterior sample in ABC, there are only few principled general-purpose approaches to the selection or construction of such summary statistics. In this paper, we develop a novel framework for this task using kernel-based distribution regression. We model the functional relationship between data distributions and the optimal choice (with respect to a loss function) of summary statistics using kernel-based distribution regression. We show that our approach can be implemented in a computationally and statistically efficient way using the random Fourier features framework for large-scale kernel learning. In addition to that, our framework shows superior performance when compared to related methods on toy and real-world problems.
\end{abstract}

\section{Introduction}
Complex generative models arise in many application domains, e.g. when we are interested in modeling population dynamics in ecology \cite{Wood:2010aa, lopes2010use}, performing phylogenetic inference and disease dynamics modeling in epidemiology \cite{Poon2015, tanaka2006using} or modeling galaxy demographics in cosmology \cite{weyant2013likelihood,cameron2012approximate}. In these models, it is often difficult or impossible to perform exact posterior inference due to an expensive to evaluate or intractable likelihood function. Approximate Bayesian computation (ABC) \cite{Beaumont02} is an inference framework that approximates the true likelihood based on the similarity between the observed and simulated data as measured by a predefined set of summary statistics. Unless the chosen summary statistics are sufficient, an information loss associated with the projection of the data onto the lower-dimensional subspace of the summary statistics occurs. This results in an approximation bias in the likelihood and subsequently in the posterior sample that is difficult to estimate. More precisely, this information loss implies that ABC performs inference on the partial posterior of the model parameters given the summary statistics of the observed data $p(\theta\vert s(y^{*}))$ in lieu of doing it on the full posterior $p(\theta\vert y^{*})$. Thus, the choice of appropriate problem-specific summary statistics is of crucial importance for the quality of posterior inference in ABC.

Several methods exist in the literature for the selection or construction of summary statistics. A number of these methods can be assembled around the idea of constructing summary statistics by linear or non-linear regression from the full dataset or a set of candidate statistics. In addition to considerations about the sufficiency of summary statistics, all of these methods require either expert knowledge for the selection of the set of candidate statistics, e.g. \citet{nakagome13}, or perform complex and high-dimensional regression by using the full dataset, e.g. \citet{Fearnhead2012}. Other examples of this approach include \citet{blum2010non}, \citet{boulesteix2007partial} and \citet{wegmann2009efficient}.

In this paper, we develop a novel framework for the construction of appropriate problem-specific summary statistics. Following the approach of \citet{Fearnhead2012}, we want to derive summary statistics that will allow inference about certain parameters of interest to be as accurate as possible. Thus, we study loss functions and reason about the optimality of summary statistics in terms of minimizing appropriate instances of these functions. In particular, we model the functional relationship between data distributions and the optimal choice of summary statistics using kernel-based distribution regression \cite{Szabo2015}.
In order to properly account for the nature of the data, we take a two-step approach to distribution regression. Additionally, we present two different variants of our framework to account for the diverse structural properties present in real-world data.

In the first variant of our framework, we assume that all aspects of the data are important for estimating the optimal summary statistics, i.e.\ we model the full marginal distribution of the data and regress from it into the space of optimal summary statistics. First, we embed the empirical distributions of newly simulated data via the mean embedding into a reproducing kernel Hilbert space (RKHS). For this embedding, we choose a characteristic kernel \cite{Sriperumbudur2011} to ensure that no information from the data is lost. We then regress from these embeddings to the optimal choice of summary statistics with kernel ridge regression \cite{friedman2001elements}. The space of candidate regressors is thus another RKHS of functions whose domain is the space of mean-embedded data distributions. For the construction of this RKHS, one can use a simple linear kernel or more flexible kernels defined on distributions such as those described in \citet{Christmann10}. The learned regression function can then be used as the summary statistics within ABC.

For the second variant of our framework, we assume that only certain aspects of the data have a direct influence on the parameter of interest in ABC and thus we restrict our attention to modeling the functional relationship between these aspects of the data and the optimal summary statistics. In particular, we assume that the observed data can be decomposed into \emph{important} and \emph{auxiliary} components such that the parameter of interest depends on the \emph{auxiliary} components of the data only through the family of induced conditional distributions of the \emph{important} components of the data given the \emph{auxiliary} ones. In order to model the functional relationship between conditional distributions and the optimal choice of summary statistics, we embed these distributions with a conditional embedding operator \cite{song2013kernel} into an RKHS and use kernel ridge regression to regress from the space of conditional embedding operators into the space of optimal summary statistics. The space of candidate regressors is thus another RKHS defined on the space of bounded linear operators between RKHS defined on the auxiliary and important components of the data, respectively.  For the construction of this RKHS, one can use a simple linear kernel or any kernel given in terms of the Hilbert-Schmidt operator norm on the difference of operators. As before, the learned regression function can be used as the summary statistics within ABC.

In this paper, we specifically study the choice of summary statistics and use a simple rejection sampling mechanism. While more complex sampling mechanisms are possible, we take this particular approach in order to decouple the influence of these two important components of ABC on the quality of posterior inference. The rest of the paper is organized as follows. Section 2 gives an overview of related work, while section 3 introduces and discusses our framework. Experimental results on toy and real-world problems, and a comparison with related methods are given in section 4. Section 5 concludes.

\section{Related Work}
Most existing methods for the selection or construction of informative summary statistics can be grouped into three categories. A first category assembles methods that first perform \emph{best subset selection} in a set of candidate statistics according to various information-theoretic criteria and then use this subset as the summary statistics within ABC. In particular, optimal subsets are selected according to, e.g. a measure of sufficiency \cite{Joyce_Marjoram_08}, entropy \cite{Nunes2010} and AIC/BIC \cite{blum2013}.

A second category consists of methods that construct summary statistics from auxiliary models. An example of this approach is indirect score ABC \cite{gleim2013approximate}. Here, a score vector that is calculated from the partial derivatives of the auxiliary likelihood plays the role of the summary statistics. Motivated by the fact that the score of the observed data is zero when the auxiliary model parameters are set by maximum-likelihood estimation (MLE), the method searches the parameter space for values whose simulated data produce a score close to zero under the same auxiliary model parameters. Thus, the discrepancy measure between the observed and simulated data is defined in terms of scores of the simulated data at the parameter values estimated with MLE from the observed data. A detailed review of this class of approaches can be found in \citet{drovandi2015}.

A third, and last, category is comprised of methods that construct summary statistics using regression from either the full dataset or a set of candidate statistics, e.g. \citet{Fearnhead2012}. \citet{Aeschbacher12} provides a general overview of such approaches, while we discuss the aforementioned method in more detail. The semi-automatic ABC (SA-ABC) method \cite{Fearnhead2012} focuses on deriving summary statistics that will allow inference about certain parameters of interest to be as accurate as possible. \citet{Fearnhead2012} focus on the construction of summary statistics that allow inference to be accurate with respect to a specific loss function. They show that the true posterior mean of the model parameters is the optimal choice of summary statistics under the quadratic loss function. As this quantity cannot be analytically calculated, they estimate it by fitting a regression model from simulated data. In particular, given simulated data $\{(\theta_{i}, y_{i})\}_{i}$, a linear model $\theta = \beta g(y) + \epsilon$ is fitted; here, $g(\cdot)$ is taken to be either the identity function or a power function. The resulting estimates $s(y) = \hat{\beta}g(y)$ are used as the summary statistics in ABC.

In the literature, there are a few other methods that are not strictly aligned with the above categorization. Here, we review three such methods -- synthetic likelihood ABC \cite{Wood:2010aa}, K-ABC \cite{nakagome13} and K2-ABC \cite{park2015k2}. Synthetic likelihood ABC \cite{Wood:2010aa} assumes that the summary statistics follow a multivariate normal distribution and uses plug-in estimates for the mean and covariance parameters. In order to generate posterior samples, the method utilizes MCMC with a synthetic likelihood that is derived by convolving the fitted distribution of the summary statistics with a Gaussian kernel that measures the similarity between the observed and simulated data via the fitted summary statistics. On the other hand, K-ABC \cite{nakagome13} and K2-ABC \cite{park2015k2} use the RKHS framework in connection with ABC, albeit in a different fashion than our framework. K-ABC regresses from already chosen summary statistics $s(y)$ to posterior expectations of interest, i.e. it estimates a conditional mean embedding operator mapping from $s(y)$ to the corresponding model parameters $\theta$. While the use of a kernel on the summary statistics increases their representative power, the method does not eliminate the challenge of selecting these summary statistics. A potential solution to this shortcoming could be to choose the whole dataset to regress with, i.e. use $s(y)=y$. This differs from our approach in two ways. First, the choice of an appropriate kernel that can be defined directly on the data is not straightforward. Our approach does not suffer from this shortcoming since we treat datasets as bags of samples. Second, instead of performing regression to estimate posterior expectations, we utilize it to calculate summary statistics that can be used within ABC. This decouples the regression model from the actual ABC method and thus, does not limit the number of samples that can be used within ABC, i.e. it allows for an arbitrary large number of samples to be drawn after performing the regression step. The K-ABC method has recently been used in HIV epidemiology \cite{Poon2015}. On the other hand, K2-ABC embeds the empirical data distributions into an RKHS via the mean embedding and uses a dissimilarity measure on the space of these embeddings to assess the similarity between the observed and simulated data. In particular, the maximum mean discrepancy is used as the dissimilarity measure on the space of the mean-embedded data distributions, and an exponential smoothing kernel is utilized to compute the ABC posterior sample weights. In contrast to the methods discussed above, there is no explicit construction or selection of summary statistics, but rather the summary statistics are given implicitly as the mean embeddings of the empirical data distributions into an possibly infinite dimensional RKHS. Our framework is different from this method in that it performs an additional step and regresses from the mean embeddings to the space of summary statistics optimal with respect to a loss function.

\section{DR-ABC Method}
In this section, we introduce and discuss the novel framework of ABC with kernel-based distribution regression (DR-ABC) and review some of its important building blocks.

\textbf{MMD.}
Given a probability distribution $F_{A}$ defined on a non-empty set $\mathcal{A}$, the mean embedding of $F_{A}$, $\mu_{F_{A}} = \mathbb{E}_{A \sim F_{A}} k(\cdot, A)$, is an element of the RKHS $\mathcal{H}_{k}$ defined by the kernel $k: \mathcal{A} \times \mathcal{A} \rightarrow \mathbb{R}$. For two probability distributions $F_{A}$ and $F_{B}$, the maximum mean discrepancy (MMD) between $F_{A}$ and $F_{B}$ is defined as
\begin{align*}
\text{MMD}^{2}(F_{A}, F_{B}) & =  \vert\vert \mu_{F_{A}} - \mu_{F_{B}} \vert\vert _{\mathcal{H}_{k}}^{2} \\ & = \mathbb{E}_{A}\mathbb{E}_{A^{\prime}} k(A, A^{\prime}) + \mathbb{E}_{B}\mathbb{E}_{B^{\prime}} k(B, B^{\prime}) \\ & \quad  - 2 \mathbb{E}_{A}\mathbb{E}_{B} k(A,B)
\end{align*}
with $A, A^{\prime}\iid F_{A}$ and $B, B^{\prime}\iid F_{B}$. Given samples $\{a_{i}\}_{i = 1}^{n_{A}}\iid F_{A}$ and $\{b_{j}\}_{j=1}^{n_{B}}\iid F_{B}$, an unbiased estimate of the MMD can be computed as
\begin{align*}
&\widehat{\text{MMD}}^{2} (F_{A}, F_{B}) = \frac{1}{n_{A}(n_{A}-1)} \sum_{i=1}^{n_{A}}\sum_{i^{\prime}\neq i} k(a_{i}, a_{i^{\prime}}) + \\ & \frac{1}{n_{B}(n_{B}-1)} \sum_{j=1}^{n_{B}}\sum_{j^{\prime}\neq j} k(b_{j}, b_{j^{\prime}}) - \frac{2}{n_{A}n_{B}} \sum_{i=1}^{n_{A}}\sum_{j=1}^{n_{B}} k(a_{i}, b_{j}).
\end{align*}

\textbf{Distribution Regression.}
The goal of distribution regression is to establish a functional relationship between probability distributions over a given set and real-valued (possibly multidimensional) responses.
In particular, given data $\{(\theta_{l}, P_{l})\}_{l=1}^{L}$ drawn i.i.d.\ from a meta distribution $\mathcal{M}$ defined on the product space of responses and probability distributions on the space of observations, we are interested in capturing this data-generating mechanism with a regression model and predicting new responses $\theta_{L+1}$ given new distributions $P_{L+1}$.

In this setting, one major challenge arises due to the fact that the probability distributions $\{P_{l}\}_{l=1}^{L}$ are not observed directly, but are available only in terms of their i.i.d. samples. In particular, the data is given as $\{(\theta_{l}, \{y_{l}^{(n)}\}_{n = 1}^{N_{l}})\}_{l=1}^{L}$ with $y_{l}^{(1)},\dots, y_{l}^{(N_{l})} \iid P_{l}$ and $\mathcal{Y}$ the underlying sample space. Thus, one is interested in predicting new responses $\theta_{L+1}$ given a new bag of samples $\{y_{L+1}^{(n)}\}_{n=1}^{N_{L+1}} \iid P_{L+1}$. This particularity makes regressing directly from the space of probability distributions $\mathcal{M}_{1}^{+}(\mathcal{Y})$ to the response space $\Theta \subset \mathbb{R}^{D}$ difficult as one has to capture the two-stage sampled nature of the data in one functional relationship.

If we take the kernel ridge regression approach, then the functional relationship between $P$ and $\theta$ is modeled as an element $g$ from the RKHS $\mathcal{G} = \mathcal{G}(k_{\mathcal{G}})$ of functions mapping from $\mathcal{M}_{1}^{+}(\mathcal{Y})$ to $\Theta$ with the kernel $k_{\mathcal{G}}$ defined on $\mathcal{M}_{1}^{+}(\mathcal{Y})$. In order to properly account for the two-stage sampled nature of the data, we take a two-stage approach to distribution regression. First, a distribution $P\in\mathcal{M}_{1}^{+}(\mathcal{Y})$ is mapped via the mean embedding $\mu$ into the RKHS $\mathcal{H}_{k}$ defined by the kernel $k:\mathcal{Y}\times\mathcal{Y} \rightarrow \mathbb{R}$. Second, this result is composed with an element $h$ from the RKHS $\mathcal{H}_{K}$ defined by the kernel $K: Y\times Y \rightarrow \mathbb{R}$, where $Y$ is the image of $\mathcal{M}_{1}^{+}(\mathcal{Y})$ under the mean embedding. Finally, this yields $k_{\mathcal{G}}(P, P^{\prime}) = K(\mu_{P}, \mu_{P^{\prime}})$ and $g = h\circ \mu$ with $h:Y \rightarrow \mathbb{R}^{D}$ such that $h(\cdot) = (h_{1}(\cdot), \dots, h_{D}(\cdot))$ and $h_{d}\in\mathcal{H}_{K}$ for every $d\in\{1, \dots, D\}$, i.e. \ we treat every dimension of $\theta$ separately. Taking the classical regularization approach, the solution of kernel ridge regression can be calculated as
\begin{equation*}
h_{d}^{\lambda} = \underset{h_{d}\in\mathcal{H}_{K}}{\argmin}\frac{1}{L} \sum_{l=1}^{L} \left\vert h_{d}\left(\mu_{\hat{P}_{l}}\right) - \theta_{ld}\right\vert ^{2} + \lambda \vert\vert h_{d}\vert\vert_{\mathcal{H}_{K}}^{2} \label{dist_reg}
\end{equation*}
with $\hat{P}_{l} = \frac{1}{N_{l}}\sum_{n=1}^{N_{l}} \delta_{y_{l}^{(n)}}$, $\theta_{l} = (\theta_{l1}, \dots, \theta_{lD})$ and $\lambda$ the regularization parameter. Given a new $P_{L+1}\in \mathcal{M}_{1}^{+}(\mathcal{Y})$ in terms of samples $\{y_{L+1}^{(n)}\}_{n=1}^{N_{L+1}}\iid P_{L+1}$, a prediction for $\theta_{L+1}$ can be calculated in the following way
\begin{equation}
\hat{\theta}_{L+1} = \boldsymbol{\Theta}(\bm{K} + L\lambda\bm{Id})^{-1}\bm{k}, \label{ridge_sol}
\end{equation}
where
$\bm{K}_{ll^{\prime}} = K(\mu_{\hat{P}_{l}}, \mu_{\hat{P}_{l^{\prime}}})$,
$\bm{k}_{l} = K(\mu_{\hat{P}_{l}}, \mu_{\hat{P}_{L+1}})$,
$\boldsymbol{\Theta} = (\theta_{1}, \dots, \theta_{L})$ and $l,l^{\prime}\in \{1, \dots, L\}$. \\

\textbf{Distribution Regression from Conditional Distributions.}
Often only certain aspects of the data are assumed to have a direct influence on the response,  especially in hierarchical or spatio-temporal modeling, and thus one might be interested in modeling only these aspects of the data. This motivates a decomposition of the data as $y_{l}^{(n)}=(z_{l}^{(n)}, x_{l}^{(n)})$ with $x_{l}^{(n)}$ encoding the \emph{important} aspects of the data (for the inference task at hand) and $z_{l}^{(n)}$ describing the rest of the information that we are not interested in modeling explicitly (e.g. this could correspond to locations on a grid where the observations are recorded).\footnote{In particular, $\left\{\left(z_{l}^{(n)},x_{l}^{(n)}\right)\right\}_{n=1}^{N_{l}}\iid P_{l}$, $\mathcal{Y} = \mathcal{Z}\times\mathcal{X}$ and $x_{l}^{(n)} \sim P_{l}(\cdot\vert z_{l}^{(n)})$.} In other words, we assume that $\theta_{l}$ depends on $P_{l}$ only through the induced conditionals $\{ P_{l}(\cdot\vert z_{l}^{(n)})\}_{n = 1}^{N_{l}}$, and thus the problem of distribution regression reduces to the question of modeling the functional relationship between the induced family of conditional distributions $\{P(\cdot\vert z)\}_{z\in\mathcal{Z}}\subset\mathcal{M}_{1}^{+}(\mathcal{X})$ and the response $\theta$, i.e. learning a map from the set of functions $\mathcal{T}=\{t:\mathcal{Z}\rightarrow \mathcal{M}_{1}^{+}(\mathcal{X}), \; t(z) = P(\cdot\vert z)\}$
into the response space $\Theta$. In order to ensure that all the necessary mathematical objects exist and are well-defined, we make the following assumptions:
\begin{enumerate}
\item $(\mathcal{X},\mathcal{B}(\mathcal{X}))$ is a Polish space with $\mathcal{B}(\mathcal{X})$ the associated Borel $\sigma$-algebra,
\item $(\mathcal{Z}, \mathscr{Z})$ is a measurable space with $\mathscr{Z}$ the associated $\sigma$-algebra,
\item kernel $k$ is bounded and can be factorized as $k((z,x), (z^{\prime}, x^{\prime})) = k_{\mathcal{Z}}(z, z^{\prime})k_{\mathcal{X}}(x, x^{\prime})$, and
\item $\mathbb{E}_{X\vert z}\left[g(X)\vert z\right]\in\mathcal{H}_{k_{\mathcal{X}}}$ for all $g\in\mathcal{H}_{k_{\mathcal{X}}}, z\in\mathcal{Z}$.
\end{enumerate}

In contrast to distribution regression from joint distributions, here we are interested in simultaneously embedding whole families of conditional distributions into an RKHS. To achieve this, we model this embedding as a function
\begin{equation*}
\mu_{X\vert \cdot} : \mathcal{Z} \rightarrow \mathcal{H}_{k_{\mathcal{X}}} \quad \text{with} \quad \mu_{X\vert Z=z} = \mathbb{E}_{X\vert Z=z} k_{\mathcal{X}}(\cdot, X),
\end{equation*}
i.e. for every $z\in\mathcal{Z}$, the embedding of the conditional distribution of $X$ given $Z=z$ is a function in the RKHS $\mathcal{H}_{k_{\mathcal{X}}}$. Following the approach of \citet{song2013kernel}, we encode the embedding of $\{P(\cdot\vert z)\}_{z\in\mathcal{Z}}$ with the conditional embedding operator $C_{X\vert Z}$, where
\vspace{-0.2cm}
\begin{equation*}
\mu_{X\vert Z=z} = C_{X\vert Z}k_{\mathcal{Z}}(\cdot, z)
\end{equation*}
and
\vspace{0.2cm}
\begin{equation*}
C_{X\vert Z} = C_{XZ}C^{-1}_{ZZ}\in \mathcal{L}(\mathcal{H}_{k_{\mathcal{Z}}}, \mathcal{H}_{k_{\mathcal{X}}}).
\end{equation*}
Thus, the embedding of a family of conditional distributions is modeled as an operator between RKHS defined on $\mathcal{Z}$ and $\mathcal{X}$, respectively. Next, we regress from the space of conditional embedding operators, i.e. \ from the space of bounded linear operators $\mathcal{L}(\mathcal{H}_{k_{\mathcal{Z}}}, \mathcal{H}_{k_{\mathcal{X}}})$, into $\Theta$ using kernel ridge regression. For this purpose, we define a kernel $K:\mathcal{L}(\mathcal{H}_{k_{\mathcal{Z}}}, \mathcal{H}_{k_{\mathcal{X}}})\times \mathcal{L}(\mathcal{H}_{k_{\mathcal{Z}}}, \mathcal{H}_{k_{\mathcal{X}}}) \rightarrow \mathbb{R}$ to measure the similarity between different conditional embedding operators. Typical choices for this kernel include the linear kernel $K(C, C^{\prime}) = Tr(CC^{\prime})$ or any other kernel given in terms of $\vert\vert C - C^{\prime}\vert\vert_{HS}$ with $\vert\vert \cdot \vert\vert_{HS}$ the Hilbert-Schmidt operator norm. Finally, putting all the building blocks together, the solution of kernel ridge regression can be computed as
\begin{equation*}
h_{d}^{\lambda} = \underset{h_{d}\in \mathcal{H}_{K}}{\argmin} \frac{1}{L} \sum_{l=1}^{L} \Big\vert h_{d}\Big(\hat{C}^{(l)}_{X\vert Z}\Big) - \theta_{ld}\Big\vert ^{2} + \lambda_{2} \vert\vert h_{d}\vert\vert_{\mathcal{H}_{K}}^{2}, \label{cond_dist_reg}
\end{equation*}
where $\hat{C}_{X\vert Z}^{(l)} = \bm{k}_{X}^{(l)}\Big(\bm{k}_{ZZ}^{(l)} + \lambda_{1} Id\Big)^{-1}\bm{k}_{Z}^{(l)}$ with
$\bm{k}_{X}^{(l)} = \left[k_{\mathcal{X}}(\cdot, x_{l}^{(1)}), \dots, k_{\mathcal{X}}(\cdot, x_{l}^{(N_{l})})\right]$,
$\left[\bm{k}^{(l)}_{ZZ}\right]_{ij} = k_\mathcal{Z}(z_{l}^{(i)}, z_{l}^{(j)})$,
$\bm{k}_{Z}^{(l)} = \left[k_{\mathcal{Z}}(\cdot, z_{l}^{(1)}), \dots, k_{\mathcal{Z}}(\cdot, z_{l}^{(N_{l})})\right]^{T}$, $\theta_{l} = (\theta_{l1}, \dots, \theta_{lD})$ and $\lambda = (\lambda_{1}, \lambda_{2})$ is the regularization parameter. Given a new distribution $P_{L+1}\in\mathcal{M}_{1}^{+}(\mathcal{Z}\times\mathcal{X})$ in terms of samples $\{z_{L+1}^{(n)}, x_{L+1}^{(n)}\}_{n=1}^{N_{L+1}}\iid P_{L+1}$ with $x_{L+1}^{(n)} \sim P_{L+1}(\cdot\vert z_{L+1}^{(n)})$, a prediction for $\theta_{L+1}$ can be calculated as
\begin{equation}
\hat{\theta}_{L+1} =
\boldsymbol{\Theta}(\bm{K} + L\lambda_{2}\bm{Id})^{-1}\bm{k}, \label{ridge_sol_cond}
\end{equation}
where
$\bm{K}_{ll^{\prime}} = K(\hat{C}_{X\vert Z}^{(l)}, \hat{C}_{X\vert Z}^{(l^{\prime})})$,
$\bm{k}_{l} = K(\hat{C}_{X\vert Z}^{(l)}, \hat{C}_{X\vert Z}^{(L+1)})$,
$\boldsymbol{\Theta} = (\theta_{1}, \dots, \theta_{L})$ and $l,l^{\prime}\in \{1, \dots, L\}$. \\

\textbf{DR-ABC.}

Our framework, ABC with kernel-based distribution regression, provides a novel approach to the construction of appropriate problem-specific summary statistics.
Motivated by \citet{Fearnhead2012}, we study loss functions and use simulated data to construct approximations to optimal summary statistics with respect to these loss functions. While any loss function can be used in our framework,\footnote{For loss functions not admitting closed-form solutions for the argument of their minimum, numerical optimization techniques might need to be used.} we focus on the quadratic loss function $L(\theta; \theta^{*}) = (\theta^{*} - \theta)^{2}$ with $\theta^{*}$ the true value of the parameter of interest. Given simulated data $\{(\theta_{l}, \{y_{l}^{(n)}\}_{n=1}^{N_{l}})\}_{l=1}^{L}$, we regress into the space of optimal summary statistics, i.e. into the parameter space in the case of quadratic loss, with kernel-based distribution regression. As discussed in the previous section, we study two different variants of our framework -- \emph{full} and \emph{conditional} DR-ABC -- to account for the diverse structural properties present in real-world data.

\underline{\emph{Full DR-ABC:}} In this variant of the DR-ABC framework, we assume that all aspects of the data are important for estimating the parameter of interest, i.e. we model the complete marginal distribution of the data and regress from it into the parameter space. In particular, we first embed the empirical distributions of the simulated data via the mean embedding into the RKHS $\mathcal{H}_{k}$ defined by the kernel $k:\mathcal{Y}\times \mathcal{Y} \rightarrow \mathbb{R}$. Next, we define a second RKHS $\mathcal{H}_{K}$ via the kernel $K: Y\times Y \rightarrow\mathbb{R}$,
\begin{equation*}
K(\mu_{\hat{P}_{l}}, \mu_{\hat{P}_{l^{\prime}}})  = \exp\Biggl(-\frac{\widehat{\text{MMD}}^{2}\left(P_{l}, P_{l^{\prime}}\right)}{2\sigma_{K}^{2}}\Biggl)
\end{equation*}
with $\sigma_{K}$ the kernel bandwidth.
This kernel provides a dissimilarity measure on the space of mean embeddings. Third, we perform kernel ridge regression from $Y$ into the parameter space with $\mathcal{H}_{K}$ as the space of candidate regressors. Finally, for a particular $P_{L+1}\in\mathcal{M}^{+}_{1}(\mathcal{Y})$ given in terms of a sample $\{y_{L+1}^{(n)}\}_{n=1}^{N_{L+1}}\iid P_{L+1}$, the approximated optimal summary statistics are equal to Equation \ref{ridge_sol}, i.e. the value of the fitted distribution regression function evaluated at the empirical distribution of that sample.

\underline{\emph{Conditional DR-ABC}}: Unlike \emph{full DR-ABC}, here we assume that only certain aspects of the data have a direct influence on the parameter of interest. Thus, we restrict our attention to modeling the functional relationship between these aspects of the data and the parameter of interest. First, we identify the \emph{important} and \emph{auxiliary} aspects of the data, i.e. we decompose the simulated data as $\{y_{l}^{(n)}\}_{n,l} = \{(z_{l}^{(n)}, x_{l}^{(n)})\}_{n,l}$. Second, for every $l$, we encode the embedding of the induced family of conditional distributions $\{P_{l}(\cdot\vert z_{l}^{(n)})\}_{n}$ with the conditional embedding operator $C_{X\vert Z}^{(l)}: \mathcal{H}_{k_{\mathcal{Z}}} \rightarrow \mathcal{H}_{k_{\mathcal{X}}}$, where $\mathcal{H}_{k_{\mathcal{Z}}}$ and $\mathcal{H}_{k_{\mathcal{X}}}$ are RKHS defined on $\mathcal{Z}$ and $\mathcal{X}$, respectively, and $k_{\mathcal{Z}}$ and $k_{\mathcal{X}}$ are the corresponding kernels. Third, we define a new RKHS $\mathcal{H}_{K}$ via the kernel $K:\mathcal{L}(\mathcal{H}_{k_{\mathcal{Z}}}, \mathcal{H}_{k_{\mathcal{X}}})\times \mathcal{L}(\mathcal{H}_{k_{\mathcal{Z}}}, \mathcal{H}_{k_{\mathcal{X}}}) \rightarrow \mathbb{R}$,
\begin{equation*}
K(C, C^{\prime}) = Tr(CC^{\prime}).
\end{equation*}
This kernel defines a dissimilarity measure on the space of conditional embedding operators. Next, we perform kernel ridge regression from this space into the parameter space and use the newly constructed RKHS $\mathcal{H}_{K}$ as the space of candidate regressors. Finally, the fitted distribution regression function can be used as a summary statistics within ABC; the approximated optimal summary statistics of a new dataset are given by Equation \ref{ridge_sol_cond}.

\underline{\emph{Application to ABC:}} Having learned a regression model, we can now perform ABC. First, we sample $M$ points from the prior and generate the corresponding datasets $\{(\theta_{m}, \{y_{m}^{(j)}\}_{j=1}^{J_{m}})\}_{m=1}^{M}$. Depending on the inference task at hand and the structural properties of the data, we may or may not perform a splitting of the data, i.e. $\{(\theta_{m}, \{(z_{m}^{(j)}, x_{m}^{(j)})\}_{j=1}^{J_{m}})\}_{m=1}^{M}$. In order to assess the similarity between the observed and simulated data, we estimate the optimal summary statistics for each dataset and compare these approximations via a smoothing
kernel that defines a dissimilarity measure on the parameter space. In particular, we calculate one of the following
\begin{align*}
& \kappa(\hat{P}_{m}, \hat{P}^{*}) = \exp\left(-\frac{\left\vert\left\vert h^{\lambda} \circ \mu_{\hat{P}_{m}} - h^{\lambda} \circ \mu_{\hat{P}^{*}}\right\vert\right\vert^{2}_{2}}{\epsilon}\right) \\ & \kappa(\hat{P}_{m}, \hat{P}^{*}) = \exp\left(-\frac{\left\vert\left\vert h^{\lambda} \circ \hat{C}_{X\vert Z}^{(m)} - h^{\lambda} \circ \hat{C}_{X\vert Z}^{*}\right\vert\right\vert^{2}_{2}}{\epsilon}\right)
\end{align*}
depending on whether we are in the setting of full or conditional DR-ABC, respectively. Here, $\hat{P}^{*}$ and $\hat{P}_{m}$, and $\hat{C}_{X\vert Z}^{*}$ and $\hat{C}_{X\vert Z}^{(m)}$ are the empirical data distributions and conditional embedding operators of the observed and newly simulated data, respectively.

Putting together kernel-based distribution regression and ABC as discussed above, the following algorithms summarize the two different variants of the DR-ABC framework.

\begin{algorithm}
   \caption{Distribution Regression}
   \label{alg:DistReg}
\begin{algorithmic}
   \STATE {\bfseries Input:} prior $\pi$ and data-generating process $P$
   \STATE {\bfseries Output:} fitted regression function $h^{\lambda} \circ \mu$
   \FOR{$l = 1, \dots, L$}
   \STATE Sample $\theta_{l} \sim \pi$
   \STATE Sample dataset $\{y_{l}^{(n)}\}_{n} \sim P(\cdot\vert \theta_{l})$
   \ENDFOR
   \STATE Fit distribution regression
   with $\{(\theta_{l}, \{y_{l}^{(n)}\}_{n})\}_{l}$
\end{algorithmic}
\end{algorithm}
\vspace{-0.2cm}

\begin{algorithm}[H]
\caption{Conditional Distribution Regression}
\label{alg:CondDistReg}
\begin{algorithmic}
\STATE {\bfseries Input:} prior $\pi$ and data-generating process $P$
\STATE {\bfseries Output:} fitted regression function $h^{\lambda} \circ C_{X\vert Z}$
     \FOR{$l = 1, \dots, L$}
           \STATE Sample $\theta_{l} \sim \pi$
           \STATE Sample dataset $\{y_{l}^{(n)}\}_{n} \sim P(\cdot\vert \theta_{l})$
           \STATE Split dataset $\{y_{l}^{(n)}\}_{n} = \{(z_{l}^{(n)}, x_{l}^{(n)})\}_{n}$
      \ENDFOR
      \STATE Fit distribution regression from conditionals
      with  \\ $\{(\theta_{l}, \{(z_{l}^{(n)}, x_{l}^{(n)})\}_{n})\}_{l}$
\end{algorithmic}
\end{algorithm}
\vspace{-0.3cm}

\begin{algorithm}
\caption{DR-ABC Algorithm}
\label{alg:DR_ABC}
\begin{algorithmic}
\STATE {\bfseries Input:} prior $\pi$, data-generating process $P$, observed \\ \hspace{1cm} data $\{y^{*}_{i}\}_{i}$, soft threshold $\epsilon$
\STATE {\bfseries Output:} weighted posterior sample $\sum_{i} w_{i} \delta_{\theta_{i}}$
\STATE {\bfseries Step 1:} Perform Distribution Regression {\bfseries \emph{or}}
 \\ \hspace{1cm} Conditional Distribution Regression depending \\ \hspace{1cm}
on the nature of the data
\STATE {\bfseries Step 2:} ABC
     \FOR{$j = 1, \dots, M$}
           \STATE Sample  $\theta_{j} \sim \pi$
           \STATE Sample dataset $\{y_{j}^{(k)}\}_{k} \sim P(\cdot\vert \theta_{j})$
           \STATE Compute $\; \tilde{w}_{j} = \exp\left(-\frac{\left\vert\left\vert h^{\lambda} \circ \mu_{\hat{P}_{j}} - h^{\lambda} \circ \mu_{\hat{P}^{*}}\right\vert\right\vert_{2}^{2}}{\epsilon}\right)$ \hspace{0.2cm} {\bfseries \emph{or}}
           \STATE \hspace{1.3cm} $\; \tilde{w}_{j} = \exp\left(-\frac{\left\vert\left\vert h^{\lambda} \circ \hat{C}_{X\vert Z}^{(j)} - h^{\lambda} \circ \hat{C}_{X\vert Z}^{*}\right\vert\right\vert_{2}^{2}}{\epsilon}\right)$
           \STATE  depending on the nature of the data
     \ENDFOR
\STATE $w_{k} = \tilde{w}_{k}/ \sum_{j = 1}^{M} \tilde{w}_{j}$ for $k = 1, \dots, M$
\end{algorithmic}
\end{algorithm}

\textbf{Computational complexity}.\ Assuming that both the size of the observed and simulated datasets is $N$, the cost of computing $\widehat{\text{MMD}}^{2}$ between two bags of samples or computing $K(\hat{C}_{X\vert Z}^{(l)}, \hat{C}_{X\vert Z}^{(l^{\prime})})$ for any $l, l^{\prime}$ is $O(N^{2})$. Given $L$ and $M$ as the number of simulated datasets for (conditional) distribution regression and ABC, respectively, the total computational cost of fitting the regression model and running ABC is $O(N^{2}(ML + L^{2}) + L^{3})$ in both \emph{full} and \emph{conditional DR-ABC}. In order to mitigate the large computational cost of our two methods, we apply the popular large-scale kernel learning framework of random Fourier features (RFF) \cite{Rahimi07randomfeatures}. This framework has successfully been applied in several contexts \cite{chitta2012efficient, huang2013random}, extended \cite{le2013fastfood,yang2014carte} and thoroughly analyzed \cite{Bach2015, sutherland2015error,sriperumbudur2015optimal}. The context most similar to ours is that of \citet{Jitkrittum2015} where two layers of
random Fourier features are applied in connection with distribution regression, albeit in the context of emulating Expectation Propagation messages.
Using random Fourier features, we approximate the potentially infinite-dimensional feature maps that figure in the computation of kernel functions with finite-dimensional vectors. This implies that kernel evaluations can be approximated by inner products of these finite-dimensional features. Using $f$ random Fourier features in each layer of approximation, we get a significantly reduced computational cost of $O(Nf(ML + L^{2}) + f^{3})$ for full DR-ABC. For conditional DR-ABC, the computational cost can be reduced to $O(f^{2}(ML + L^{2}) + f^{3})$. In our experiments, we use $f = 100$ and the following RFF expansion
\begin{equation*}
\hat{\phi}(x) \in\mathbb{R}^{f}, \quad \hat{\phi}(x)_{i:(i+1)} = \sqrt{\frac{2}{f}} [\cos(w_{i}^{T}x), \sin(w_{i}^{T}x)].
\end{equation*}

Due to a result from \citet{Bach2015}, a comparatively small number of random Fourier features can be used even for large datasets since the number of random Fourier features needed for good approximations of kernel ridge regression solutions often scales sublinearly with the number of observations. Nevertheless, the dependence of the required number of random Fourier features on the number of datasets and the number of observations within each dataset, particularly in settings such as ours where there are two layers of random Fourier features, is not yet fully understood.

\section{Experimental Results}
{\bfseries Toy example.} The first problem we study is the following Gaussian hierarchical model
\begin{align*}
\theta &\sim \mathcal{N}(2,1), \\
z &\sim \mathcal{N}(0,2), \\
x\vert z, \theta &\sim\mathcal{N}(\theta z^{2},1).
\end{align*}
This simple example serves as a proof of concept for our framework. In this model, the parameter of interest is $\theta$, and our goal is to estimate $\mathbb{E}[\theta\vert y^{*}]$ with $y^{*}$ the observed dataset. In our experiments, we compare our two methods, full and conditional DR-ABC, against SA-ABC and K2-ABC. We specifically compare our framework against these methods as they are examples illustrating the regression and RKHS approach to the construction of summary statistics. For the performance metric, we calculate the mean square error (MSE) of the parameter of interest on synthetic data. In particular, we set $\theta^{*} = 2$ and generate $200$ observations given this parameter value as $y^{*}$; for every newly simulated dataset, we also draw $200$ datapoints. For full DR-ABC, we take the kernels $k$ and $K$ as Gaussian kernels, while for conditional DR-ABC, $k$ is a Gaussian kernel and $K$ is a linear kernel. The hyperparameters in the two DR-ABC methods are set via
five-fold cross-validation on appropriately defined grids. For the grids of the different kernel bandwidth parameters, we multiply the respective median heuristics \cite{reddi2014decreasing}
with a set of ten equally spaced points between $10^{-4}$ and $1000$. For $\lambda$ and $\epsilon$, we choose the grids by exponentiating $10$ to the powers given by ten equally spaced points between $-4$ and $1$. In order to account for the randomness in the generative process, we run each of the methods $20$ times and display the mean of MSE across the different runs.

\begin{figure}[ht]
\vskip -0.15in
\begin{center}
\centerline{\includegraphics[width=\columnwidth]{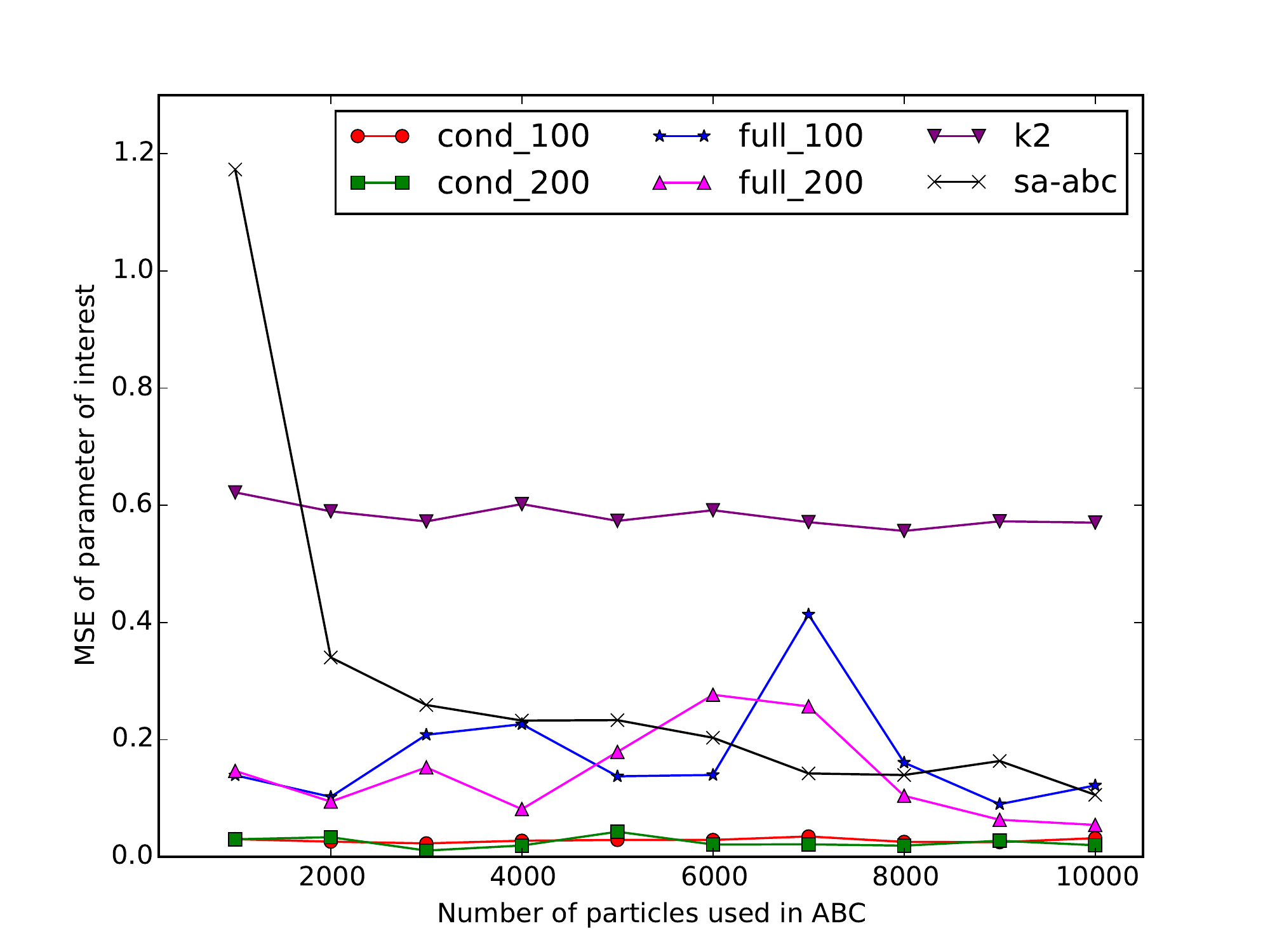}}
\caption{MSE of the parameter of interest for full and conditional DR-ABC, SA-ABC and K2-ABC averaged across $20$ runs. The number of ABC particles is between 1000 -- 10000. We use either $100$ or $200$ particles in (conditional) distribution regression.}
\label{fig:toy}
\end{center}
\vskip -0.2in
\end{figure}

Figure \ref{fig:toy} describes the performance of our chosen methods across different numbers of particles used in the (conditional) distribution regression phase (for full and conditional DR-ABC) and in the ABC phase (for all four methods). In order to achieve comparable results, we use the same number of particles in the regression phase as in the ABC phase  for SA-ABC.

K2-ABC exhibits a fairly stable reconstruction error for different numbers of ABC particles and outperforms SA-ABC only when relatively small numbers of ABC particles are used. Across the wide spectrum of the number of ABC particles, we see both conditional and full DR-ABC outperforming K2-ABC by a large margin. While full DR-ABC usually also outperforms SA-ABC, conditional DR-ABC does this consistently by a large margin.

{\bfseries Ecological Dynamical Systems.} Many ecological problems have an intractable likelihood due to a dynamic generative process and thus rely on ABC for posterior inference. Deriving appropriate summary statistics in this setting is quite challenging as the dependence structure of the data needs to be appropriately taken into account. As an example of an ecological system with a dynamic generative process, we examine the problem of inferring the dynamics of the adult blowfly population as introduced in \citet{Wood:2010aa}. In particular, the population dynamics are modeled by the following discretised differential equation
\begin{equation*}
N_{t+1} = PN_{t-\tau}\exp\left(-\frac{N_{t-\tau}}{N_{0}}\right)e_{t} + N_{t}\exp(-\delta\epsilon_{t})
\end{equation*}
with $N_{t+1}$ denoting the observation at time $t+1$ which is determined by the time-lagged observations $N_{t}$ and $N_{t-\tau}$, and the Gamma distributed noise variables $e_{t}\sim$ Gam$(\frac{1}{\sigma_{p}^{2}}, \sigma_{p}^{2})$ and $\epsilon_{t}\sim$ Gam$(\frac{1}{\sigma_{d}^{2}}, \sigma_{d}^{2})$. The parameters of interest in this model are $\theta=\{P, N_{0}, \sigma_{d}, \sigma_{p}, \tau, \delta\}$. As before, we compare our framework with SA-ABC and K2-ABC. The performance metric and the kernel and hyperparameter selection is done in the same way as in the previous example.

\begin{figure}[H]
\vskip -0.1in
\begin{center}
\centerline{\includegraphics[width=\columnwidth]{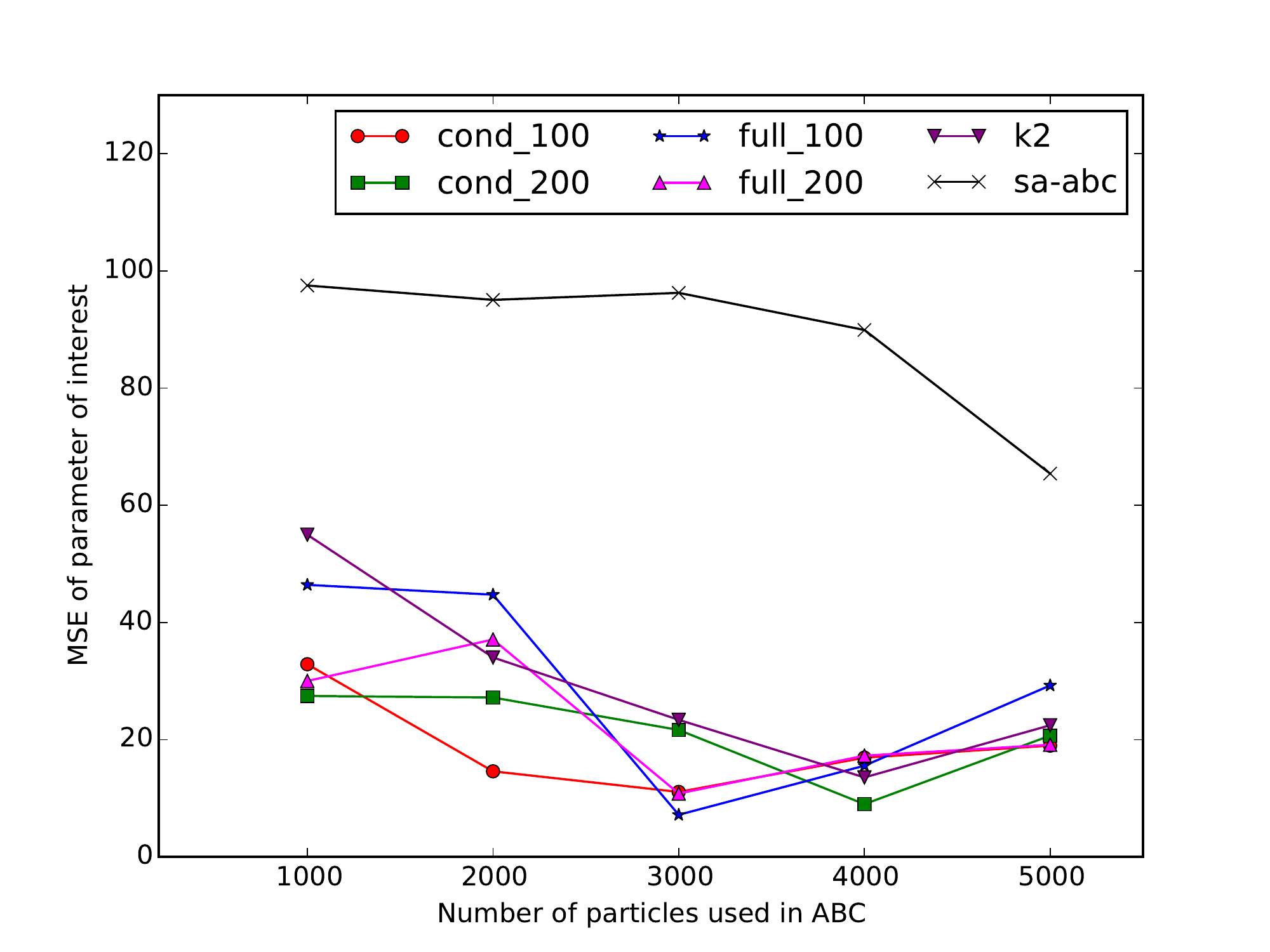}}
\caption{MSE of the parameter of interest for full and conditional DR-ABC, SA-ABC and K2-ABC averaged across $20$ runs. The number of ABC particles is between 1000 - 5000. We use either $100$ or $200$ particles in (conditional) distribution regression.}
\label{fig:blowfly}
\end{center}
\vskip -0.2in
\end{figure}

From Figure \ref{fig:blowfly}, we that our methods outperform SA-ABC by a large margin across the whole spectrum of test situations. Full DR-ABC displays competitive performance to K2-ABC, even outperforming it in certain instances by a large margin. On the other hand, conditional DR-ABC outperforms K2-ABC in all test situations; in some of these situations, the performance of our method is massively superior.

{\bfseries Lotka-Volterra Model.} Another popular ecological model in which posterior inference is difficult is the Lotka-Volterra model \cite{alfred1925lotka,volterra1927variazioni}. This model describes the dynamics of biological systems in which two species interact in a predator-prey relationship.
\begin{align*}
&\frac{dx}{dt} = \alpha x - \beta xy, \\
&\frac{dy}{dt} = \gamma xy - \delta y,
\end{align*}
where $x,y$ are the number of prey and predators, respectively, $\alpha, \beta, \gamma, \delta$ are positive real parameters describing the interaction of the two species, $t$ denotes time and $\frac{dx}{dt}$,$\frac{dy}{dt}$ are the respective growth rates. In addition to the dynamical nature of the generative process, the interaction between the two species makes deriving informative summary statistics even more challenging. The parameters of interest in this model are $\theta = \{\alpha, \beta, \gamma, \delta \}$. As in the previous two experiments, we compare our framework with SA-ABC and K2-ABC and use the same performance metric, kernels and hyperparameter selection procedure.

\begin{figure}[ht]
\begin{center}
\centerline{\includegraphics[width=\columnwidth]{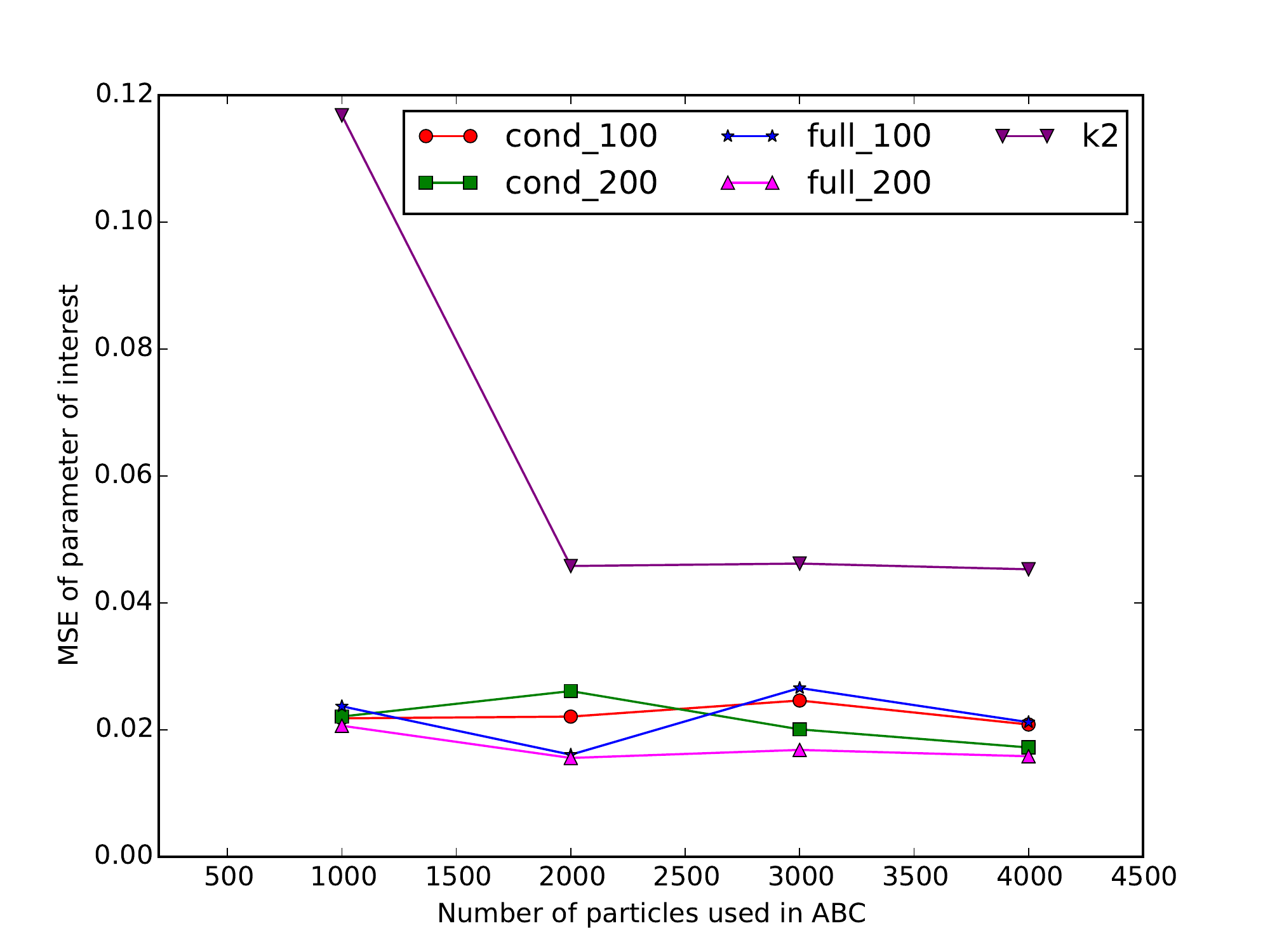}}
\caption{MSE of the parameter of interest for full and conditional DR-ABC, and K2-ABC averaged across $20$ runs. The number of ABC particles is between 1000 - 4000.  We use either $100$ or $200$ particles in (conditional) distribution regression.}
\label{fig:lv}
\end{center}
\vskip -0.2in
\end{figure}

From Figure \ref{fig:lv}, we see that our framework outperforms competing methods by a large margin. While for small numbers of ABC particles, full DR-ABC seems to perform better, for large numbers of ABC particles, conditional DR-ABC slightly outperforms full DR-ABC with a clear downward trend in the error for higher numbers of ABC particles. As for SA-ABC, the method cannot directly be applied to this problem due to the high correlation in the observations which leads to a regression problem that is ill conditioned. In order to mitigate this, we performed principal component analysis and used an approximation of the data matrix given by the first $10$ principal components in SA-ABC. The resulting errors are 1--2 orders of magnitude larger than those displayed in the figure and are thus excluded from it for clarity.

\section{Conclusion}
In this paper, we developed a novel framework for the construction of informative problem-specific summary statistics using the flexible and expressive setting of reproducing kernel Hilbert spaces. We introduced two different approaches based on embeddings of probability distributions and kernel-based distribution regression. Our proposed framework has several advantages over previous general-purpose and semi-automatic summary statistics construction methods. First, by using the flexible RKHS framework, we are able to regulate the kind and amount of information that is extracted from the data
and thus construct more informative problem-specific summary statistics, as opposed to mandating an ad hoc selection of a limited set of candidate statistics or postulating heuristic summary statistics which inevitably leads to a hard to evaluate approximation bias in the likelihood and subsequently in the posterior sample. Moreover, our framework compactly encodes the extracted information. Second, due to the modeling flexibility of our framework, we are able to appropriately account for different structural properties present in real-world data. Third, our methods can be implemented in a computationally and statistically efficient way using the random Fourier features framework for large-scale kernel learning. Fourth, our framework can be easily extended to any object class on which the embedding kernel(s) can be defined. Examples of such object classes include genetic data \cite{Wu2010}, phylogenetic trees \cite{Poon2015},
strings, graphs and other structured data \cite{Gaertner2003}. Fifth, although there are multiple sets of hyperparameters in each of our methods, their selection can be performed in a principled way via cross-validation.
From experimental results on toy and real-world problems, we see that our framework substantially reduces the bias in the posterior sample achieving superior performance when compared to related methods that used for the construction of summary statistics in ABC.

\bibliography{abc_biblio,dr-abc_biblio}
\bibliographystyle{icml2016}

\end{document}